\documentclass[cmbright]{mmaauth}

\usepackage{moreverb}

\usepackage[dvips,colorlinks,bookmarksopen,bookmarksnumbered,
             citecolor=red,urlcolor=red]{hyperref}

\usepackage{graphicx}        
\usepackage{multicol}        
\usepackage[bottom]{footmisc}
\usepackage{bm}
\usepackage{amsmath}
\usepackage{amssymb}
\usepackage{caption}
\usepackage[noadjust]{cite}

 \newtheorem{thm}{Theorem}

 \newtheorem{rmk}[thm]{Remark}

 \numberwithin{equation}{section}

       \newcommand{\R}{\ensuremath{\mathbb{R}}}
       \newcommand{\N}{\ensuremath{\mathbb{N}}}

       \newcommand{\HQ}{\ensuremath{\mathbb{H}}}
       
       \newcommand{\be}{\begin{equation}}
       \newcommand{\ee}{\end{equation}}
       
       \newcommand{\qi}{\ensuremath{\mbox{\boldmath $i$}}}
       
       \newcommand{\qj}{\ensuremath{\mbox{\boldmath $j$}}}
       
       \newcommand{\qk}{\ensuremath{\mbox{\boldmath $k$}}}
       
       \newcommand{\vect}[1]{\mbox{\boldmath $#1$}}
       \newcommand{\bv}[1]{\mathrm{\mathbf{#1}}}
       \newcommand{\eprior}{\ensuremath{\underline{e}}}
       \newcommand{\epost}{\ensuremath{\overline{e}}}

\newcommand\BibTeX{{\rmfamily B\kern-.05em \textsc{i\kern-.025em b}\kern-.08em
T\kern-.1667em\lower.7ex\hbox{E}\kern-.125emX}}

\def\volumeyear{2011}

\begin{document}

\runninghead{E.~Hitzer}

\title{Algebraic foundations of split hypercomplex nonlinear adaptive filtering}

\author{E.~Hitzer\corrauth}

\address{Department of Applied Physics, University of Fukui, 3-9-1 Bunkyo, 910-8507 Fukui, Japan}

\corraddr{Department of Applied Physics, University of Fukui, 3-9-1 Bunkyo, 910-8507 Fukui, Japan}

\begin{abstract}
A split hypercomplex learning algorithm for the training of nonlinear finite impulse response adaptive filters for the processing of hypercomplex signals of any dimension is proposed. The derivation strictly takes into account the laws of hypercomplex algebra and hypercomplex calculus, some of which have been neglected in existing learning approaches (e.g. for quaternions). Already in the case of quaternions we can predict improvements in performance of hypercomplex processes. The convergence of the proposed algorithms is rigorously analyzed.
\end{abstract}

\MOS{60G35; 15A66}

\keywords{Quaternionic adaptive filtering,
Hypercomplex adaptive filtering,
Nonlinear adaptive filtering,
Hypercomplex Multilayer Perceptron,
Clifford geometric algebra}

\maketitle

\vspace{-6pt}

\section{Introduction}

Split quaternion nonlinear adaptive filtering has recently been treated by \cite{UTM:SQNAF}, who showed its superior performance for Saito's Chaotic Signal and for wind forecasting. 
The quaternionic methods constitute a generalization of complex valued adaptive filters, treated in detail in \cite{MG:CVNAF}. 
A method of quaternionic least mean square algorithm for adaptive quaternionic has previously been developed in \cite{UM:QLMS}. 
Additionally, \cite{UTM:QVNAF} successfully proposes the usage of \textit{local analytic} fully quaternionic functions in the Quaternion Nonlinear Gradient Descent (QNGD). 
Yet the unconditioned use of analytic fully quaternionic activation functions in neural networks faces problems with poles due to the Liouville theorem \cite{SB:thesis}. 
The quaternion algebra of Hamilton is a special case of the higher dimensional Clifford algebras \cite{HS:CAtoGC}. The problem with poles in nonlinear analytic functions does not generally occur for hypercomplex activation functions in Clifford algebras, where the Dirac and the Cauchy-Riemann operators are not elliptic \cite{FS:private}, as shown, e.g., for hyperbolic numbers in \cite{EH:HypAF}. But in order to provide a general approach valid for all Clifford algebras, we will only use \textit{split} hypercomplex nonlinear functions. 

Our present paper first retraces some of the work of \cite{UTM:SQNAF} in order to show modifications due to the strict application of quaternionic calculus (Section \ref{sc:SQNF}).  We then introduce the wider framework of hypercomplex algebra\footnote{Hypercomplex numbers include, but are not limited to: real, complex, hyperbolic numbers, quaternions, Segre quaternions, Grassmann numbers, real and complex vectors, Lie algebras, Lie algebra of oriented spheres, multivector algebras, geometric algebras, Clifford algebras, commutative hypercomplex numbers, Pauli algebra, Dirac algebra, spacetime algebra, versor algebras, and spin algebras.} and hypercomplex calculus in Section \ref{sc:HypNum}, where hypercomplex refers to Clifford geometric algebra. Hypercomplex calculus is also known as geometric calculus or Clifford analysis and has been thoroughly presented, including hypercomplex differentiation and integration, in \cite{HS:CAtoGC} and \cite{EH:MVDC}. Finally, in Sections \ref{sc:SHAFA} and \ref{sc:AASHAFA} we generalize split quaternion nonlinear adaptive filtering to split \textit{hypercomplex} nonlinear adaptive filtering. We thus remove the dimensional limitations due to the use of quaternions, and introduce the algebraically much richer framework of hypercomplex numbers. Our general treatment is not limited to division algebras \cite{BTH:OptLR, BHT:Coindupd}. 

Hypercomplex numbers allow to describe a wide variety of geometric objects (points, lines, planes, hyperplanes, circles, spheres, hyperspheres) by elementary algebraic entities, their transformations and interactions by elementary algebraic operations (\cite{HTBY:Carrier}). For the interested reader additional information about the products of hypercomplex numbers and their geometric interpretation are summarized in Appendix \ref{ap:geoint}. The use of hypercomplex numbers for neural networks allows to directly learn these objects and their transformations in the form of hypercomplex numbers (\cite{EH:GopsGANN}). To a certain degree complex and quaternion neural networks can fulfill this task in lower dimensions (\cite{TN:CVNN}). 
Hypercomplex neural networks, including the universal approximation properties of split hypercomplex activation functions, have been thoroughly studied by \cite{SB:thesis}. \cite{FMX:AttCtrl} applies a quaternionic multilayer perceptron (neural network) approach and quaternionic radial basis functions to rigid body attitude control. Especially the question of optimal learning rates for hypercomplex neural networks, not limited to division algebras, have been studied in \cite{BTH:OptLR, BHT:Coindupd}. 

In the current research we extend hypercomplex networks to split hypercomplex nonlinear FIR filtering, and for signals with large dynamical ranges to adaptive amplitude split hypercomplex nonlinear FIR filtering. Due to the algebraic complexity, which needs to be duly taken care of before actual numerical computations become possible, this paper is concentrating on presenting the algebraic foundations in a selfcontained (as far as space allows) and unified way.

\section{Split quaternion nonlinear adaptive filtering \label{sc:SQNF}}

\subsection{Split quaternion nonlinear functions in \HQ}

Gauss, Rodrigues and Hamilton's four-dimensional (4D) quaternion algebra $\HQ$ is defined over $\R$ 
with three imaginary units:

\be
 \qi \qj = -\qj \qi = \qk, \,\,
 \qj \qk = -\qk \qj = \qi, \,\,
 \qk \qi = -\qi \qk = \qj, \,\, 
 \qi^2=\qj^2=\qk^2=\qi \qj \qk = -1.
\label{eq:quat}
\end{equation}
Every quaternion can be written explicitly as
\be
  q=q_r + q_i \qi + q_j \qj + q_k \qk \in \HQ, \quad 
  q_r,q_i, q_j, q_k \in \R,
  \label{eq:aquat}
\end{equation}
and has a \textit{quaternion conjugate}\footnote{Quaternion conjugation is equivalent to \textit{reversion} in $Cl_{3,0}^+$, and to \textit{principal involution} in $Cl_{0,2}$.}
\be
  \tilde{q} = q_r - q_i \qi - q_j \qj - q_k \qk.
\end{equation}
This leads to the \textit{norm} of $q\in\HQ$
\be
  | q | = \sqrt{q\tilde{q}} = \sqrt{q_r^2+q_i^2+q_j^2+q_k^2},
  \qquad
  | p q | = | p || q |.
\end{equation}
The \textit{scalar} part of a quaternion is defined as
\be
  Sc(q) = q_r = \frac{1}{2}(q+\tilde{q}),
  \label{eq:Hscal}
\ee
the \textit{pure} quaternion (non-scalar) part is
\be 
  \vect{q} = q - q_r = q_i \qi + q_j \qj + q_k \qk = \frac{1}{2}(q-\tilde{q}).
  \label{eq:qvect}
\ee 
The coefficients of \vect{q} can be extracted by
\be 
  q_i = -\frac{1}{2}(\vect{q}\qi+\qi\vect{q}), \,\,
  q_j = -\frac{1}{2}(\vect{q}\qj+\qj\vect{q}), \,\,
  q_k = -\frac{1}{2}(\vect{q}\qk+\qk\vect{q}).
  \label{eq:purecomp}
\ee  
The \textit{product} of two quaternions $w,x \in \HQ$ can be expanded with \eqref{eq:quat} as
\begin{align}
  wx = & \,(w_r + w_i \qi + w_j \qj + w_k \qk)(x_r + x_i \qi + x_j \qj + x_k \qk)\nonumber\\
     =  & \,w_rx_r - w_ix_i - w_jx_j - w_kx_k     \nonumber\\
     \phantom{=} & \,+(w_rx_i + w_ix_r + w_jx_k - w_kx_j)\qi \nonumber\\
     \phantom{=} & \,+(w_rx_j + w_jx_r + w_kx_i - w_ix_k)\qj \nonumber\\
     \phantom{=} & \,+(w_rx_k + w_kx_r + w_ix_j - w_jx_i)\qk.
  \label{eq:wxprod}
\end{align}
\begin{rmk}
\label{rm:quatprod}
Note that one full quaternion product requires the computation of 16 real multiplications and 12 real additions. 
\end{rmk}

A \textit{split} quaternion nonlinear function is a real analytic and bounded nonlinearity $\phi: \R \rightarrow \R$ applied independently to each component of the quaternion-valued signal. 
\be 
  \Phi(q) 
  = \phi_r(q) + \phi_i(q) \qi + \phi_j(q) \qj + \phi_k(q) \qk,
\ee 
with 
\be 
  \phi_r(q)=\phi(q_r), \,\,
  \phi_i(q)=\phi(q_i), \,\,
  \phi_j(q)=\phi(q_j), \,\,
  \phi_k(q)=\phi(q_k).
  \label{eq:splitphi}
\ee
Therefore each function $\phi_r, \phi_i, \phi_j, \phi_k$ is a nested function obtained by applying \eqref{eq:qvect}, followed by \eqref{eq:purecomp}, followed by $\phi$. This approach is not analytic in \HQ, but it is componentwise analytic and bounded, suitable for neural networks.

\subsection{Quaternionic nonlinear adaptive filtering}

\begin{figure}
\begin{center}
\includegraphics[width=10cm]{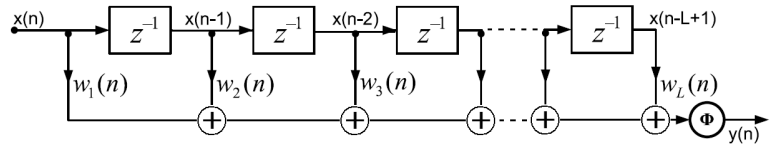}
\end{center}
\caption{Schematic diagram of a nonlinear adaptive finite impulse response (FIR) filter. \label{fg:FIR}}
\end{figure}

We now study a quaternionic learning algorithm for nonlinear adaptive finite impulse response (FIR) filters, compare Fig. \ref{fg:FIR}.
The \textit{input vector} $\bv{x}=\bv{x}_n$ has $L\in \N$ quaternion valued components 
$x_n, \ldots , x_{n-L+1} \in \HQ$. Similarly the adaptive \textit{weight vector} $\bv{w}=\bv{w}_n$ has $L\in \N$ quaternion valued components 
$w_1, \ldots , w_{L} \in \HQ$. Each of these vectors can also be written in terms of four $L$-dimensional real vectors as 
$\bv{x} = \bv{x}_r + \bv{x}_i \qi + \bv{x}_j \qj +\bv{x}_k \qk, 
\,\,\bv{x}_r, \bv{x}_i, \bv{x}_j, \bv{x}_k \in \R^L $, and 
$\bv{w} = \bv{w}_r + \bv{w}_i \qi + \bv{w}_j \qj +\bv{w}_k \qk, 
\,\,\bv{w}_r, \bv{w}_i, \bv{w}_j, \bv{w}_k \in \R^L $, respectively. The \textit{net input} $s_n\in\HQ$ is given by the inner product
\be 
  s_n = \bv{w}\cdot\bv{x} = \sum_{l=1}^L w_l \,x_{n-l+1}.
  \label{eq:expip}
\ee 
Expanding the quaternion products $w_l \,x_{n-l+1}$ according to \eqref{eq:wxprod} we get the four real coefficients of $s_n$ as 
\begin{align}
  s_{n,r} &= \bv{w}_r\cdot \bv{x}_r - \bv{w}_i\cdot \bv{x}_i - \bv{w}_j\cdot \bv{x}_j - \bv{w}_k\cdot \bv{x}_k     \nonumber\\
    s_{n,i} &= \bv{w}_r\cdot \bv{x}_i + \bv{w}_i\cdot \bv{x}_r + \bv{w}_j\cdot \bv{x}_k - \bv{w}_k\cdot \bv{x}_j \nonumber\\
    s_{n,j} &= \bv{w}_r\cdot \bv{x}_j + \bv{w}_j\cdot \bv{x}_r + \bv{w}_k\cdot \bv{x}_i - \bv{w}_i\cdot \bv{x}_k \nonumber\\
    s_{n,k} &= \bv{w}_r\cdot \bv{x}_k + \bv{w}_k\cdot \bv{x}_r + \bv{w}_i\cdot \bv{x}_j - \bv{w}_j\cdot \bv{x}_i
  \label{eq:sexp}
\end{align}

The application of a split quaternion nonlinear function $\Phi$ to the net input signal $s_n$ leads to the quaternionic FIR \textit{output signal} $y_n \in \HQ$
\begin{align} 
  y_n &= y_{n,r} + y_{n,i} \qi + y_{n,j} \qj + y_{n,k} \qk \nonumber \\
      &= \Phi(s_n) 
      = \phi_r(s_{n}) + \phi_i(s_{n})\qi + \phi_j(s_{n})\qj + \phi_k(s_{n})\qk.
  \label{eq:yn}
\end{align} 

\cite{UTM:SQNAF} define in equation (A.3) the \textit{quaternionic} vector derivative $\partial_{\bv{w}}$ as
\be 
  \partial_{\bv{w}_r} y_{n} 
  + \partial_{\bv{w}_i} y_{n}\qi 
  + \partial_{\bv{w}_j} y_{n}\qj
  + \partial_{\bv{w}_k} y_{n}\qk
  = y_n \partial_{\bv{w}},   
  \label{eq:defdw}
\ee 
where we understand $\partial_{\bv{w}}$ as acting from the \textit{right} on the function $y_n = \Phi(s_n) = \Phi(\bv{w}\cdot \bv{x})$. The reason for this order is, that the quaternion products in definition \eqref{eq:defdw} do not commute, e.g. $y_{n}\qi \neq \qi y_{n}$, i.e. the product order matters. For clarity note that $\partial_{\bv{w}_i} y_{n}\qi = (\partial_{\bv{w}_i} y_{n})\qi$, where $\partial_{\bv{w}_i}$ is a real $L$-dimensional vector derivative. Similarly $\partial_{\bv{w}_r}$, $\partial_{\bv{w}_j}$, $\partial_{\bv{w}_k}$ are real $L$-dimensional vector derivatives. 

The difference of the desired quaternionic FIR \textit{target signal} $d_n\in \HQ$ and the output $y_n$ yields the quaternionic \textit{error} 
\be 
  e_n = d_n - y_n = e_{n,r} + e_{n,i} \qi + e_{n,j} \qj + e_{n,k} \qk. 
\ee 
The real quaternionic FIR filter cost function is defined as
\begin{align} 
  E_n &= | e_n|^2 = e_n \widetilde{ e_n} = e_{n,r}^2 + e_{n,i}^2 + e_{n,j}^2 + e_{n,k}^2
      = (d_n - y_n)(d_n - y_n)^{\sim} \nonumber \\
      &= (d_{n,r} - y_{n,r})^2 + (d_{n,i} - y_{n,i})^2 
      + (d_{n,j} - y_{n,j})^2 + (d_{n,k} - y_{n,k})^2. 
\end{align}

\begin{rmk} \label{rm:dwnonc} 
  The general \textit{algebraic} non-commutativity of the quaternionic vector derivative operator $\partial_{\bv{w}}$ with quaternion functions cannot be emphasized enough, since it is a \textit{characteristic distinction} of quaternion calculus (and hypercomplex calculus in general) from real and complex calculus. For example in general 
$E_n \partial_{\bv{w}} 
  = (e_n \widetilde{e_n})\partial_{\bv{w}}
  \neq (e_n\partial_{\bv{w}})\widetilde{e_n} 
       + e_n(\widetilde{e_n}\partial_{\bv{w}})$, see \cite{HS:CAtoGC}, a fact which may have been neglected in (14), (17) and (20) of \cite{UTM:SQNAF}.
\end{rmk}

\subsection{Learning algorithm for split quaternionic FIR nonlinear adaptive filters}

The aim of a quaternion-valued learning algorithm for nonlinear filtering is to iteratively minimize $E_n$ (\cite{SO:LMSalg}). \textit{Gradient descent update} leads\footnote{\cite{UTM:SQNAF}, Section 3.1, do not have the factor $1/2$ of \eqref{eq:graddesc}, but e.g. \cite{UTM:SQNAF}, equation (16), implies this factor $1/2$ without further explanation.} to
\be 
  \bv{w}_{n+1} = \bv{w}_n - \frac{1}{2} \mu\, E_n \partial_{\bv{w}},
  \label{eq:graddesc}
\ee 
where $\mu \in \R$ is the \textit{learning rate} (or \textit{step size}). We therefore need to compute
\begin{align} 
  E_n \partial_{\bv{w}} 
      &=\phantom{+}  
          e_{n,r}^2\partial_{\bv{w}} 
        + e_{n,i}^2\partial_{\bv{w}} 
        + e_{n,j}^2\partial_{\bv{w}} 
        + e_{n,k}^2\partial_{\bv{w}} \nonumber \\
      &= -2e_{n,r}\,y_{n,r}\partial_{\bv{w}} 
        - 2 e_{n,i}\,y_{n,i}\partial_{\bv{w}} 
        - 2e_{n,j}\,y_{n,j}\partial_{\bv{w}} 
        - 2e_{n,k}\,y_{n,k}\partial_{\bv{w}}, 
  \label{eq:Edw}
\end{align}
where we used 
\be 
  e_{n,r}^2\partial_{\bv{w}} 
  = 2e_{n,r}(e_{n,r}\partial_{\bv{w}})
  = 2e_{n,r}(d_{n,r}\partial_{\bv{w}}-y_{n,r}\partial_{\bv{w}})
  = -2e_{n,r}(y_{n,r}\partial_{\bv{w}}),
\ee 
because 
$d_{n,r}\partial_{\bv{w}}=0$, and the analogous equations obtained by index substitution $r \rightarrow i,j,k$. 

We will first compute 
\begin{align}
  y_{n,r}\partial_{\bv{w}} 
   &= \phi_r(s_n) \partial_{\bv{w}} 
    = \phi(s_{n,r})\partial_{\bv{w}} \nonumber \\
    &= \partial_{\bv{w}_r} \phi(s_{n,r}) 
  + \partial_{\bv{w}_i} \phi(s_{n,r})\qi 
  + \partial_{\bv{w}_j} \phi(s_{n,r})\qj
  + \partial_{\bv{w}_k} \phi(s_{n,r})\qk \nonumber \\
  &= \phi'(s_{n,r})[\partial_{\bv{w}_r}s_{n,r}
                    + \partial_{\bv{w}_i}s_{n,r} \qi 
                    + \partial_{\bv{w}_j}s_{n,r} \qj 
                    + \partial_{\bv{w}_k}s_{n,r} \qk]\nonumber \\
  &= \phi'(s_{n,r})[\bv{x}_r-\bv{x}_i\qi-\bv{x}_j\qj-\bv{x}_k\qk]
  \nonumber \\
  &= \phi'_r(s_{n}) \widetilde{\bv{x}}, 
  \label{eq:ynrdw}
\end{align}
where we used \eqref{eq:yn} for the first equality, \eqref{eq:splitphi} for the second equality, \eqref{eq:defdw} for the third equality, the chain rule $\partial_{\bv{w}_r} \phi(f(\bv{w}_r)) 
= \left[\frac{\partial \phi(\lambda)}{\partial \lambda}\right]_{\lambda = f(\bv{w}_r)} \partial_{\bv{w}_r}f(\bv{w}_r)$, and $\phi'(f)=\left[\frac{\partial \phi(\lambda)}{\partial \lambda}\right]_{\lambda = f} $, for the fourth equality, \eqref{eq:sexp} and the real $L$-D vector derivative of the inner product $\partial_{\bv{w}_r}(\bv{w}_r\cdot\bv{x}_r) = \bv{x}_r$ for the fifth equality, and we defined $\phi'_r(s_{n}) = \phi'(s_{n,r})$.

Next we compute
\begin{align}
  y_{n,i}\partial_{\bv{w}} 
   &= \phi_i(s_n) \partial_{\bv{w}} 
    = \phi(s_{n,i})\partial_{\bv{w}} \nonumber \\
    &= \partial_{\bv{w}_r} \phi(s_{n,i}) 
  + \partial_{\bv{w}_i} \phi(s_{n,i})\qi 
  + \partial_{\bv{w}_j} \phi(s_{n,i})\qj
  + \partial_{\bv{w}_k} \phi(s_{n,i})\qk \nonumber \\
  &= \phi'(s_{n,i})[\partial_{\bv{w}_r}s_{n,i}
                    + \partial_{\bv{w}_i}s_{n,i} \qi 
                    + \partial_{\bv{w}_j}s_{n,i} \qj 
                    + \partial_{\bv{w}_k}s_{n,i} \qk]\nonumber \\
  &= \phi'(s_{n,i})[\bv{x}_i+\bv{x}_r\qi+\bv{x}_k\qj-\bv{x}_j\qk]
   = \phi'(s_{n,i})\qi[-\bv{x}_i\qi+\bv{x}_r-\bv{x}_k\qk-\bv{x}_j\qj]
  \nonumber \\
  &= \phi'_i(s_{n}) \,\qi\,\widetilde{\bv{x}}, 
  \label{eq:ynidw}
\end{align}
where the computations are analogous to \eqref{eq:ynrdw}, except that we need \eqref{eq:quat} for pulling out $\qi$ to the left in the sixth equality and we defined $\phi'_i(s_{n}) = \phi'(s_{n,i})$. In analogy to \eqref{eq:ynrdw} and \eqref{eq:ynidw}, and by defining $\phi'_j(s_{n}) = \phi'(s_{n,j})$, $\phi'_k(s_{n}) = \phi'(s_{n,k})$, we can derive 
\be 
  y_{n,j}\partial_{\bv{w}} = \phi'_j(s_{n}) \,\qj\,\widetilde{\bv{x}}, 
  \qquad
  y_{n,k}\partial_{\bv{w}} = \phi'_k(s_{n}) \,\qk\,\widetilde{\bv{x}}. 
  \label{eq:ynjkdw}
\ee 

Equations \eqref{eq:ynrdw}, \eqref{eq:ynidw} and \eqref{eq:ynjkdw} yield for the quaternionic weight vector derivative of the output
\begin{align} 
 y_n \partial_{\bv{w}} 
  &= 
  (y_{n,r} + \qi y_{n,i}  + \qj  y_{n,j} + \qk y_{n,k} ) \partial_{\bv{w}}
  \nonumber \\
  &= [\phi'_r(s_{n}) +\qi^2 \phi'_i(s_{n}) +\qj^2 \phi'_j(s_{n})+\qk^2 \phi'_k(s_{n})]\,\widetilde{\bv{x}}
  \nonumber \\
  &= [\phi'_r(s_{n}) - \phi'_i(s_{n}) - \phi'_j(s_{n}) - \phi'_k(s_{n})]\, \widetilde{\bv{x}}.
  \label{eq:yndw}
\end{align} 

\begin{rmk} \label{rm:phiprime}
We went to great length and detail to show \eqref{eq:yndw}, because in equation (A.8) of \cite{UTM:SQNAF}, it seems to have been wrongly assumed that $\phi'_r(s_{n}) = \phi'_i(s_{n}) = \phi'_j(s_{n}) = \phi'_k(s_{n})$. Yet our analysis clearly shows, that it is indeed the same function $\phi'$ evaluated at \textit{different} arguments $s_{n,r}$, $s_{n,i}$, $s_{n,j}$, and $s_{n,k}$!
\end{rmk}

For obtaining the quaternionic vector derivative of the cost function we insert 
\eqref{eq:ynrdw}, \eqref{eq:ynidw} and \eqref{eq:ynjkdw} in \eqref{eq:Edw} 
\be 
  E_n\partial_{\bv{w}} 
  = -2[e_{n,r}\phi'_r(s_{n}) 
      + e_{n,i}\qi\phi'_i(s_{n}) 
      + e_{n,j}\qj\phi'_j(s_{n}) 
      + e_{n,k}\qk\phi'_k(s_{n})]\, \widetilde{\bv{x}}.
  \label{eq:Edwfin}
\ee 
\begin{rmk}
The result \eqref{eq:Edwfin} for the quaternionic vector derivative of the cost function $E_n\partial_{\bv{w}}$ should be compared with (13) and (21) of \cite{UTM:SQNAF}. In our derivation of \eqref{eq:Edwfin} the non-commutativity of the quaternion product\footnote{\cite{UTM:SQNAF} claim in their derivation in Section 3.1, that they obtain the different result (21) from (13) of (\cite{AFMX:NNinMD}) by considering non-commutativity: ``However, if the non-commutativity of the quaternion product is considered as in our proposed algorithms, the error gradient becomes ..." (\cite{UTM:SQNAF}, p. 428).} was fully taken into account, yet our results are clearly different even from (21) of \cite{UTM:SQNAF}. The reasons have been pointed out in Remarks \ref{rm:dwnonc} and \ref{rm:phiprime}.
\end{rmk}

For better comparison with \cite{UTM:SQNAF}, we also compute analogous to \eqref{eq:yndw} the quaternionic vector derivative of the quaternion conjugate output $\widetilde{y_n}$ as follows\footnote{Once again we obtain a different result from equation (18) in \cite{UTM:SQNAF}, because $\phi'$ has to be evaluated with \textit{different} arguments $s_{n,r}$, $s_{n,i}$, $s_{n,j}$, and $s_{n,k}$!}
\begin{align}
  \widetilde{y_n}\partial_{\bv{w}}
  &= (y_{n,r} - \qi y_{n,i}  - \qj  y_{n,j} - \qk y_{n,k} ) \partial_{\bv{w}}
  \nonumber \\
  &= [\phi'_r(s_{n}) -\qi^2 \phi'_i(s_{n}) -\qj^2 \phi'_j(s_{n})-\qk^2 \phi'_k(s_{n})]\,\widetilde{\bv{x}}
  \nonumber \\
  &= [\phi'_r(s_{n}) + \phi'_i(s_{n}) + \phi'_j(s_{n}) + \phi'_k(s_{n})]\, \widetilde{\bv{x}}.
  \label{eq:qcyndw}
\end{align} 

Finally inserting \eqref{eq:Edwfin} into the weight update formula \eqref{eq:graddesc} we obtain the \textit{split quaternion adaptive filtering algorithm (SQAFA) weight update} as
\be 
  \bv{w}_{n+1} = \bv{w}_n + \mu\, [e_{n,r}\phi'_r(s_{n}) 
      + e_{n,i}\phi'_i(s_{n})\qi 
      + e_{n,j}\phi'_j(s_{n})\qj 
      + e_{n,k}\phi'_k(s_{n})\qk]\, \widetilde{\bv{x}}.
  \label{eq:wtupdate}
\ee 
\begin{rmk}
Note that our expression in \eqref{eq:wtupdate} is indeed less complex and easier to compute than (22) of \cite{UTM:SQNAF}. Our update $\Delta\bv{w}_n=\bv{w}_{n+1}-\bv{w}_n$ needs (apart from the common real $\mu$ factor) only four real multiplications, three additions and one full quaternion product (see Remark \ref{rm:quatprod}), compared to four full quaternion products and one full quaternion addition in (22) of \cite{UTM:SQNAF}. 
\end{rmk}

Real world signals often have large dynamical ranges. One approach to cope with the large signal dynamics is to use a \textit{trainable amplitude} for the non-linear activation function, improving performance compared to algorithms with fixed non-linearities (\cite{HM:adapamp, ET:trainamp}). 
The derivation of the \textit{adaptive amplitude} split quaternion adaptive filtering algorithm (AASQAFA) for signals of large dynamic range of \cite{UTM:SQNAF}, Section 3.3, remains valid in our approach as well. For $A\in \{r,i,j,k\}$, the nonlinear functions $\phi_A(s_n)$ are replaced by unit amplitude nonlinearities 
$\lambda_{n,A}\varphi_A(s_n)$ with trainable amplitude parameters $\lambda_{n,A}\in \R$. The amplitude updates are computed by
\be 
  \lambda_{n+1,A} 
  = \lambda_{n,A} - \frac{1}{2}\rho \frac{\partial E_n}{\partial \lambda_{n,A}}
  = \lambda_{n,A}+ \rho e_{n,A} \varphi_A(s_n),
  \quad \forall A \in \{r,i,j,k\}.
\ee

\subsection{Convergence analysis for SQAFA and AASQAFA\label{sc:Hconv}}

In \textit{convergence analysis }the relationship of the 
\textit{a posteriori} error 
\be
\epost_n = d_n -\Phi(\bv{w}_{n+1}\cdot\bv{x}_n),
\ee 
with the \textit{a priori} error 
\be 
\eprior_n = d_n -\Phi(\bv{w}_{n}\cdot\bv{x}_n),
\ee 
given by the first order Taylor expansion
\be 
  |\epost_n|^2 = |\eprior_n|^2 + (|\eprior_n|^2\partial_{\bv{w}})\cdot\widetilde{\Delta\bv{w}_n},
  \label{eq:taylor}
\ee 
is considered. It involves the quaternion conjugate of the
weight update 
\begin{align}
  \widetilde{\Delta\bv{w}_n}
  &= (\bv{w}_{n+1}-\bv{w}_n)^{\sim} 
  \nonumber \\
  &= \mu\, \bv{x} \,[e_{n,r}\phi'_r(s_{n}) 
      - e_{n,i}\phi'_i(s_{n})\qi 
      - e_{n,j}\phi'_j(s_{n})\qj 
      - e_{n,k}\phi'_k(s_{n})\qk]
  \label{eq:conjwtupdate}
\end{align}
of \eqref{eq:wtupdate}, and the error gradient $|\eprior_n|^2\partial_{\bv{w}}$ of \eqref{eq:Edwfin}. For convergence of the SQAFA, we need
$|\epost_n|^2 < |\eprior_n|^2$ under the assumptions of small learning rates $\mu$.

Inserting \eqref{eq:Edwfin} and \eqref{eq:conjwtupdate}, the second term of the Taylor expansion \eqref{eq:taylor} gives
\begin{align}
  &(|\eprior_n|^2\partial_{\bv{w}})\cdot\widetilde{\Delta\bv{w}_n}
  \nonumber \\
  &= -2\mu\,[\eprior_{n,r}\phi'_r(s_{n}) 
      + \eprior_{n,i}\phi'_i(s_{n})\qi 
      + \eprior_{n,j}\phi'_j(s_{n})\qj 
      + \eprior_{n,k}\phi'_k(s_{n})\qk]\, 
  \nonumber \\
  &\phantom{==}(\widetilde{\bv{x}} \cdot \bv{x}) 
      \,[\eprior_{n,r}\phi'_r(s_{n}) 
      - \eprior_{n,i}\phi'_i(s_{n})\qi 
      - \eprior_{n,j}\phi'_j(s_{n})\qj 
      - \eprior_{n,k}\phi'_k(s_{n})\qk]
  \\
  &= -2\mu\,(\widetilde{\bv{x}} \cdot \bv{x})
     [(\eprior_{n,r}\phi'_r(s_{n}))^2 
      + (\eprior_{n,i}\phi'_i(s_{n}))^2
      + (\eprior_{n,j}\phi'_j(s_{n}))^2
      + (\eprior_{n,k}\phi'_k(s_{n}))^2].
  \nonumber 
\end{align}
The Taylor expansion \eqref{eq:taylor} can thus be expressed as
\be
  |\epost_n|^2 = |\eprior_n|^2 [1-M],
  \label{eq:eerel}
\ee
with
\begin{align} 
   M &= 2\mu\,(\widetilde{\bv{x}} \cdot \bv{x})
    \frac{(\eprior_{n,r}\phi'_r(s_{n}))^2 
      + (\eprior_{n,i}\phi'_i(s_{n}))^2
      + (\eprior_{n,j}\phi'_j(s_{n}))^2
      + (\eprior_{n,k}\phi'_k(s_{n}))^2}{|\eprior_n|^2}
  \nonumber \\
  &\leq 2\mu\,(\widetilde{\bv{x}} \cdot \bv{x}) |P|^2,
  \label{eq:defM}
\end{align}
where we applied a 4D Cauchy-Schwarz identity to obtain
\begin{gather} 
  (\eprior_{n,r}\phi'_r(s_{n}))^2 
      + (\eprior_{n,i}\phi'_i(s_{n}))^2
      + (\eprior_{n,j}\phi'_j(s_{n}))^2
      + (\eprior_{n,k}\phi'_k(s_{n}))^2
  \leq |\eprior_n|^2 |P|^2,
  \nonumber \\
  \text{with} \quad P = \phi'_r(s_{n}) 
      + \phi'_i(s_{n}) \qi
      + \phi'_j(s_{n}) \qj
      + \phi'_k(s_{n}) \qk.
\end{gather}
For convergence we need $0<M<1$ in \eqref{eq:eerel}. 
Therefore we can ensure convergence by demanding 
\be 
  0 < \mu < \frac{1}{ 2(\widetilde{\bv{x}} \cdot \bv{x}) |P|^2}.
  \label{eq:mulimit}
\ee 
\begin{rmk}
  The upper limit for $\mu$ in \eqref{eq:mulimit} is by a factor \textit{five} higher than the stability limit given in Section 4.1 of \cite{UTM:SQNAF}! This should lead to much faster convergence.
\end{rmk}

Because in the AASQAFA the a posteriori and a priori errors, respectively, are analyzed componentwise, the results in \cite{UTM:SQNAF}, Section 4.2, continue to remain valid.

\section{Hypercomplex numbers in Clifford geometric algebras\label{sc:HypNum}}

\subsection{Clifford geometric algebras $Cl_{p,q}$ over vector spaces $\R^{p,q}$}


We now introduce hypercomplex algebras $Cl_{p,q}$ of W.K. Clifford, and develop the hypercomplex versions of SQAFA and AASQAFA, appropriately called SHAFA and AASHAFA, where the letter ``H'' stands for \textit{hypercomplex}. As a motivation observe that we can rename the quaternion units to 
$e_1 = \qi, e_2 = \qj, \,e_{12} = e_1e_2 = \qk$ and express a quaternion as a hypercomplex $Cl_{0,2}$ number\footnote{
The conventional use of $q$ as zero or positive integer index in the Clifford algebra notation $Cl_{p,q}$, and the use of $q$ as a general quaternion element in Section \ref{sc:SQNF} and (\ref{eq:qCl02}) should not be confused. The use of the same letter is somewhat unfortunate, but $q$ for quaternions is only used in (\ref{eq:qCl02}) to provide the connection to the treatment of Section \ref{sc:SQNF}.
}
\be 
  q = q_0 1 + q_1 e_1 + q_2 e_2 + q_{12}  e_{12},
  \label{eq:qCl02}
\ee 
where the Clifford algebra $Cl_{0,2}$ is the geometric algebra of the normed vector space $\R^{0,2}$ with orthonormal basis $\{e_1, e_2\}$.

We now first define the general notion of a \textit{Clifford geometric algebra} in plain mathematical terms (\cite{FM:ICNAAM2007, HS:CAtoGC}).
Let $\{e_1, e_2, \ldots , e_q, e_{q+1}, \ldots , e_n \}$, with $n=p+q$, $e_k^2=\varepsilon_k$, $\varepsilon_k = +1$ for $k=1, \ldots , q$, $\varepsilon_k = -1$ for $k=q+1, \ldots , n$,  be an orthonormal basis of the normed vector space $\R^{p,q}$ with a product according to the multiplication rules 
\be
  e_k e_l + e_l e_k = 2 \varepsilon_k \delta_{k,l}, 
  \qquad k,l = 1, \ldots n,
\label{eq:mrules}
\ee
where $\delta_{k,l}$ is the Kronecker symbol with $\delta_{k,l}= 1$ for $k=l$, and $\delta_{k,l}= 0$ for $k\neq l$. This non-commutative product generates the associative $2^n$-dimensional Clifford geometric algebra $Cl(\R^{p,q})=Cl(p,q)=Cl_{p,q} = \mathcal{G}_{p,q}$ over $\R$. 
The set\footnote{Note the font distinction between $e_k$ for basis vectors and $\vect{e}_{A_s}$ for general basis elements of the Clifford algebra, which are products of basis vectors.} $\{ \vect{e}_{A_s}: A_s\subseteq \{1, \ldots ,n\}\}$ with $\vect{e}_{A_s} = e_{h_1}e_{h_2}\ldots e_{h_r}$, $1 \leq h_1< \ldots < h_r \leq n$, $\vect{e}_{\emptyset}=1$, forms a graded basis of $Cl_{p,q}$. The grades $r$ range from $0$ for scalars, $1$ for vectors, $2$ for bivectors, $k$ for $k$-vectors, up to $n$ for so called pseudoscalars. 

The $r$-vector subspaces are spanned by the induced bases 
\be 
  \label{eq:rvecbasis}
  \{{e}_{k_1} {e}_{k_2} \ldots  {e}_{k_r}
   \mid 1 \leq k_1 < k_2 < \ldots < k_r \leq n \},
\ee
each with dimension $\binom{r}{n}$. The total dimension of the $Cl_{p,q}$ therefore becomes $\sum_{r=0}^n \binom{r}{n} = 2^n$.

The real vector space $\R^{p,q}$ will be embedded in $Cl_{p,q}$ by identifying the element $(a_1,a_2,\ldots,a_n)\in\R^n$ with the element $a=a_1e_1+a_2e_2+\ldots a_ne_n$ of the algebra. The general elements of $Cl_{p,q}$ are real linear combinations of basis blades $\vect{e}_{A_s}$, and are called Clifford numbers, multivectors or hypercomplex numbers. 

As an example we take the Clifford geometric algebra $Cl_{3}=Cl_{3,0}$ of three-dimensional (3D) Euclidean space $\R^3=\R^{3,0}$.
$\R^3$ has an orthonormal basis $\{{e}_1, {e}_2, {e}_3\}$. 
$Cl_{3}$ then has a $2^3=8$-dimensional basis of  
\be
  \label{eq:G3basis}
  \{{1}, 
    \underbrace{{e}_1, {e}_2, {e}_3}_{\text{vectors}},
    {\underbrace{{e}_2{e}_3,{e}_3{e}_1, {e}_1{e}_2}_{\text{area bivectors}}},  
    \underbrace{i={e}_1{e}_2{e}_3}_{\text{volume trivector}}\}.
\ee
Here $i$ denotes the unit trivector, i.e. the oriented volume of a unit cube, with $i^2=-1$. 
The even grade subalgebra $Cl_{3}^+$ is
isomorphic to Hamilton's quaternions $\HQ$.\footnote{%
%
As noted earlier, another Clifford algebra isomorphic to $\HQ$ is $Cl_{0,2}$ with basis 
$\{1, {e}_1, {e}_2, \vect{e}_{12}={e}_1{e}_2\}$.} 
Moreover, e.g. the subalgebra of $Cl_3$ with basis $\{1,e_1\}$ is isomorphic to hyperbolic numbers, and e.g. the subalgebras with bases $\{1,e_{12}\}$, and $\{1,e_{123}\}$ are isomorphic to complex numbers. 
For ease of notation a \textit{lexical basis order} like in \eqref{eq:G3basis} is assumed and the $Cl_{p,q}$ basis elements are indexed consecutively with $1 \leq A \leq 2^n$.

The $k$-vector parts ($0\leq k \leq n$) of a multivector
  $M \in Cl_{p,q}, p+q=n,$ can be extracted with the grade selector bracket
$\langle \ldots \rangle_k$:
  \textit{scalar} part
  $Sc(M) = \langle M \rangle = \langle M \rangle_0 = M_0 \in \R$, 
  \textit{vector} part
  $\langle M \rangle_1 \in \R^{p,q}$, 
  \textit{bi-vector} part
  $\langle M \rangle_2 \in\bigwedge^2\R^{p,q}$,  \ldots, 
  and
  \textit{pseudoscalar} part $\langle M \rangle_n\in\bigwedge^n\R^{p,q}$
\begin{equation}\label{eq:MVgrades}
    M  =  \sum_{A=1}^{2^n} M_{A} \vect{e}_{A}
       =  \langle M \rangle + \langle M \rangle_1 + \langle M \rangle_2 + \ldots +\langle M \rangle_n \, .
\end{equation}

The \textit{principal involution}\footnote{Applying no involution, or replacing the principal involution, e.g., by the reverse 
$
  {e}_{k_1} {e}_{k_2} \ldots {e}_{k_r}
  \mapsto {e}_{k_r} \ldots {e}_{k_2} \,{e}_{k_1}
$,
$
    1 \leq k_1 < k_2 < \ldots < k_r \leq n
$,
without sign changes leads in the case of $Cl_{1,0}$ to the norm expression for hyperbolic numbers, instead of \eqref{eq:Clmodulus}.}
\be 
  {}^{\sim}: Cl_{p,q} \rightarrow Cl_{p,q}, 
  \qquad
  MN \mapsto \widetilde{N}\widetilde{M},
  \label{eq:pinv}
\ee
changes the sign of all basis vectors according to ${e}_{k} \mapsto \varepsilon_k {e}_{k}$, $1\leq k \leq n$, 
and reverses the order of all vector factors
$
  {e}_{k_1} {e}_{k_2} \ldots {e}_{k_r}
  \mapsto (\varepsilon_{k_r}{e}_{k_r}) \ldots  (\varepsilon_{k_2}{e}_{k_2}) \,(\varepsilon_{k_1}{e}_{k_1})
$,
$
    1 \leq k_1 < k_2 < \ldots < k_r \leq n.
$
By linearity it extends to all multivectors $\forall M\in Cl_{p,q}: M \mapsto \widetilde{M}$, and it replaces \textit{complex conjugation} and \textit{quaternion conjugation}.
In particular $\widetilde{\alpha} = \alpha, \forall \alpha \in \R$, scalars are therefore invariant under the principal involution
\be 
  \widetilde{\langle M \rangle}  
  = \langle\, \widetilde{M} \,\,\rangle
  = \langle M \rangle.  
  \label{eq:pinvscalar}
\ee
The principal involution has the \textit{unique property} that
\be 
  \vect{e}_A \,\widetilde{\vect{e}_A} = \widetilde{\vect{e}_A}\,\vect{e}_A = 1, 
  \qquad 
  \vect{e}_A \ast \widetilde{\vect{e}_B} = \delta_{A,B}, 
  \qquad  1 \leq A,B \leq 2^n.
  \label{eq:Clbasis_scp}
\ee 

The \textit{scalar product} of two multivectors $M, {N} \in Cl_{p,q}$ is defined as
\be
    M \ast {N} 
    = \langle M{N} \rangle 
    = \langle M{N} \rangle_0.
\ee
The scalar product is symmetric 
$\langle M{N} \rangle= M \ast {N} = N \ast M =\langle NM \rangle $.
Applying the principal involution to one factor we get 
$M\ast \widetilde{N}=\sum_{A} M_A N_A$. The scalar product allows therefore to compute the $2^n$ multivector coefficients directly
\be 
  M_A = M \ast \widetilde{\vect{e}_A} = \langle M \widetilde{\vect{e}_A} \,\rangle.
  \label{eq:Clcomp}
\ee 
This corresponds to \eqref{eq:Hscal} and \eqref{eq:purecomp} for the case of quaternions with $\vect{e}_1 = 1, \vect{e}_2 = \qi, \vect{e}_3=\qj, \vect{e}_4 = \qk$.
  The \textit{modulus} $|M|$ of a multivector $M \in Cl_{p,q}$ is defined\footnote{If the principal involution in \eqref{eq:Clmodulus} is omitted, then we instead obtain $|M|^2 = M \ast M = \sum_{A}  M_A^2 e_A^2$, 
with $e_A^2 \in \{-1, +1\}$, which is useful for implementing hyperbolic numbers and their higher dimensional generalizations.} as 
\be
  |M|^2 
  = \langle M \widetilde{M}\rangle 
  = {M\ast\widetilde{M}} 
  = {\sum_{A} M_A^2}.
  \label{eq:Clmodulus}
\ee

In the subsequent discussion of the convergence conditions for hypercomplex nonlinear adaptive filtering the following hypercomplex Cauchy-Schwarz type inequality proves useful. The proof is given in \ref{ap:HCSin}.

\begin{thm}[Hypercomplex Cauchy-Schwarz type inequality]
\label{th:HCSine}
Con\-sider two general multivectors $M,N \in Cl_{p,q}$. The following inequality holds
\be 
  |M \ast N| \leq |M| |N|.
\ee 
\end{thm}

\subsection{Split hypercomplex nonlinear functions in $Cl_{p,q}$\label{sc:hypsplit}}

The \textit{Clifford product} of two multivectors $w,x \in Cl_{p,q}$ is bilinear and its $2^n$ coefficients can be again computed from
\begin{align} 
  ( wx )_A 
  = \langle wx\, \widetilde{\vect{e}_{A}}\rangle 
  = \langle \vect{e}_{A}\, \widetilde{x} \widetilde{w}\rangle 
  = \sum_{B,C=1}^{2^n} w_{B} x_{C}
    \langle \vect{e}_{A} \widetilde{\vect{e}_{C}}\,\widetilde{\vect{e}_{B}}\rangle,
  \label{eq:Cliffwx}
\end{align}
where we applied \eqref{eq:Clcomp} for the first equality, and \eqref{eq:pinv} and \eqref{eq:pinvscalar} for the second equality. Note that the real scalar coefficients in \eqref{eq:Cliffwx} commute $w_{B} x_{C}= x_{C} w_{B}$. 
For quaternions $\HQ$, $Cl_3^+$, or $Cl_{0,2}$ this will result in exactly the same $2^2=4$ bilinear coefficients as on the right side of the second equality in \eqref{eq:wxprod}. The importance of \eqref{eq:Cliffwx} lies in the fact that it is very compact, dimension independent, and allows the efficient computation of hypercomplex weight vector derivatives (compare \eqref{eq:ymAdw}).

A \textit{multivector valued function} (\cite{EH:MVDC}) 
  $f: Cl_{p,q} \rightarrow Cl_{p,q}$, has $2^n$ blade components
  $(f_A: Cl_{p,q} \rightarrow \R)$
  \begin{equation}\label{eq:MVfunc}
  f: x \mapsto 
    f(x)  =  \sum_{A=1}^{2^n} f_{A}(x) {\vect{e}}_{A},
  \qquad \forall x \in Cl_{p,q}.
  \end{equation}

A \textit{split hypercomplex nonlinear function} is a real analytic and bounded nonlinearity $\phi: \R \rightarrow \R$ applied independently to each component of the multivector-valued signal $x\in Cl_{p,q}$. 
\be 
  \Phi(x) 
  = \sum_{A=1}^{2^n}\phi_A(x){\vect{e}}_{A},
  \qquad 
  \phi_A(x)=\phi(x_A).
\ee 
Therefore each function $\phi_A, 1\leq A \leq 2^n$ is a \textit{nested} function obtained by first applying \eqref{eq:Clcomp} followed by $\phi$. This approach is not analytic\footnote{Even for complex numbers, Liouville's theorem 
states that every bounded complex analytic function is constant. 
This makes non-linear bounded complex analytic functions impossible.%
} in $Cl_{p,q}$, but it is componentwise analytic and bounded, suitable for neural networks.

\setcounter{footnote}{0}

\section{Hypercomplex nonlinear adaptive filtering \label{sc:SHAFA}}

We now study a \textit{hypercomplex} learning algorithm for nonlinear adaptive finite impulse response (FIR) filters. 
The \textit{input vector}\footnote{To avoid confusion of vector space dimension $n=p+q$ and the length of the input vector, we rename the latter now $m$.} $\bv{x}=\bv{x}_m$ has $L\in \N$ multivector valued components 
$x_m, \ldots , x_{m-L+1} \in Cl_{p,q}$. Similarly the adaptive \textit{weight vector} $\bv{w}=\bv{w}_m$ has $L\in \N$ multivector valued components 
$w_1, \ldots , w_{L} \in Cl_{p,q}$. Each of these vectors can be written in terms of $2^n$ $L$-dimensional real vectors as 
$\bv{x} = \sum_{A=1}^{2^n}\bv{x}_A\vect{e}_{A}, 
\,\,\forall A: \bv{x}_A \in \R^L $, and 
$\bv{w} = \sum_{A=1}^{2^n}\bv{w}_A\vect{e}_{A}, 
\,\,\forall A: \bv{w}_A \in \R^L $, 
respectively. The \textit{net input} $s_m\in Cl_{p,q}$ is given by the 
inner product\footnote{The \textit{inner product} maps two vectors
$\bv{x},\bv{y}$, each with $L$ multivector valued components, to a single hypercomplex number in $Cl_{p,q}$.}
\be 
  s_m = \bv{w}\cdot\bv{x} = \sum_{l=1}^L w_l \,x_{m-l+1}.
 \label{eq:Clexpip}
\ee 
According to \eqref{eq:Cliffwx} we get the $2^n$ real coefficients of $s_m$ 
as\footnote{The inner product applied to the vectors $\bv{w}_B, \bv{x}_C \in \R^L$ is indeed the standard inner product of $\R^L$, mapping pairs of vectors to real scalars.} 
\be 
  s_{m,A}
  =(\bv{w}\cdot\bv{x} )_A 
  = \sum_{B,C=1}^{2^n} (\bv{w}_{B}\cdot \bv{x}_{C})\,  
     \langle \vect{e}_{A} \widetilde{\vect{e}_{C}}\widetilde{\vect{e}_{B}}\rangle.
  \label{eq:smA}
\ee 

The application of a split hypercomplex nonlinear function $\Phi$ to the net input signal $s_m$ leads to the hypercomplex FIR \textit{output signal} $y_m \in Cl_{p,q}$
\begin{align} 
  y_m = \sum_{A=1}^{2^n}y_{m,A}\vect{e}_{A} 
      = \Phi(s_m) = \sum_{A=1}^{2^n}\phi_A(s_{m}) \vect{e}_{A}.
  \label{eq:ynCl}
\end{align} 

We now define the \textit{hypercomplex} vector derivative $\partial_{\bv{w}}$ (acting from the right\footnote{We keep the order of applying the \textit{hypercomplex} vector derivative from the right for ease of comparison with \cite{UTM:SQNAF} and our \eqref{eq:defdw}. The whole formalism can easily be established for a left derivative of the form 
$\partial_{\bv{w}}y_m 
  = \sum_{A=1}^{2^n}\vect{e}_{A}\partial_{\bv{w}_A} y_{m}$.}) 
as (\cite{EH:MVDC})
\be 
  y_m \partial_{\bv{w}}
  = \sum_{A=1}^{2^n}\partial_{\bv{w}_A} y_{m}\vect{e}_{A},   
  \label{eq:defdwCl}
\ee 
For clarity note that 
$\partial_{\bv{w}_A} y_{m}\vect{e}_{A} = (\partial_{\bv{w}_A} y_{m})\vect{e}_{A}$, $1\leq A \leq 2^n$,
where the $\partial_{\bv{w}_A}$ are the real $L$-dimensional vector derivatives.  

The difference of the desired hypercomplex FIR \textit{target signal} $d_m\in Cl_{p,q}$ and the output $y_m$ yields the \textit{error} 
\be 
  e_m = \sum_{A=1}^{2^n}e_{m,A}\vect{e}_{A}
      = d_m - y_m. 
\ee 
The real hypercomplex FIR filter \textit{cost function} is defined as
\begin{align} 
  E_m &=  e_m \ast\widetilde{ e_m} 
       = \sum_{A=1}^{2^n}e_{m,A}^2
       = (d_m - y_m)\ast(d_m - y_m)^{\sim} \nonumber \\
      &= \sum_{A=1}^{2^n}(d_{m,A}-y_{m,A})^2. 
\end{align}

\subsection{Learning algorithm for split hypercomplex FIR nonlinear adaptive filters}

The aim of a hypercomplex-valued learning algorithm for nonlinear filtering is to iteratively minimize $E_m$. \textit{Gradient descent update} leads to
\be 
  \bv{w}_{m+1} = \bv{w}_m - \frac{1}{2} \mu E_m \partial_{\bv{w}},
  \label{eq:Clgraddesc}
\ee 
with learning rate (step size) $\mu \in \R$. We therefore compute
\be 
  E_m \partial_{\bv{w}}
  = \sum_{A=1}^{2^n} e_{m,A}^2\partial_{\bv{w}}
  = -2\sum_{A=1}^{2^n} e_{m,A}(y_{m,A}\partial_{\bv{w}}),
\ee 
because
\begin{align} 
  e_{m,A}^2\partial_{\bv{w}}
  &= 2 e_{m,A}(e_{m,A}\partial_{\bv{w}})
  = 2 e_{m,A}(d_{m,A}\partial_{\bv{w}}-y_{m,A}\partial_{\bv{w}})
  \nonumber \\
  &= -2 e_{m,A}(y_{m,A}\partial_{\bv{w}}),
  \label{eq:emAdw}
\end{align} 
as $d_{m,A}\partial_{\bv{w}}=0$.

Now follows the computation of the mathematical \textit{key result} for split hypercomplex gradient descent update learning. 
The hypercomplex vector derivative of the hypercomplex FIR output signal gives
\begin{align} 
  y_{m,A}\partial_{\bv{w}}
  &= \phi_A(s_m)\partial_{\bv{w}}
  = \phi(s_{m,A})\partial_{\bv{w}}
  \nonumber \\
  &= \phi'(s_{m,A})[s_{m,A}\partial_{\bv{w}}]
  = \phi'(s_{m,A})\sum_{D=1}^{2^n}\partial_{\bv{w}_D}s_{m,A}\vect{e}_{D}
  \nonumber \\
  &= \phi'(s_{m,A})\sum_{B,C,D=1}^{2^n}\partial_{\bv{w}_D}
     (\bv{w}_{B}\cdot \bv{x}_{C})\,
     \langle \vect{e}_{A} 
     \widetilde{\vect{e}_{C}}\,\widetilde{\vect{e}_{B}}\rangle 
     \vect{e}_{D}
  \nonumber \\
  &= \phi'(s_{m,A})\sum_{B,C,D=1}^{2^n}
     \delta_{D,B}\,\bv{x}_{C}\,
     \langle \vect{e}_{A} 
     \widetilde{\vect{e}_{C}}\,\widetilde{\vect{e}_{B}}\rangle 
     \vect{e}_{D}
  \nonumber \\
  &= \phi'(s_{m,A})\sum_{B=1}^{2^n}
     \langle \vect{e}_{A} 
     \widetilde{\bv{x}}\,\widetilde{\vect{e}_{B}}\rangle 
     \vect{e}_{B}
   = \phi'(s_{m,A})\sum_{B=1}^{2^n}
     (\vect{e}_{A} \widetilde{\bv{x}})_B  \,\vect{e}_{B}
  \nonumber \\
  &=  \phi'(s_{m,A}) \,\vect{e}_{A} \widetilde{\bv{x}},
  \label{eq:ymAdw}
\end{align} 
where we inserted $s_{m,A}$ of \eqref{eq:smA} for the fifth equality, 
used 
$\partial_{\bv{w}_D}(\bv{w}_{B}\cdot \bv{x}_{C})=\delta_{D,B}\,\bv{x}_{C}$
for the sixth equality, \eqref{eq:Clcomp} for the eighth equality with coefficient
$M_B=(\vect{e}_{A} \widetilde{\bv{x}})_B$, and finally $M=\vect{e}_{A} \widetilde{\bv{x}}$. 

\begin{rmk}
  The \textit{central result} \eqref{eq:ymAdw} is very powerful, because it is valid for all Clifford algebras $Cl_{p,q}$. In particular it subsumes the quaternionic vector derivatives \eqref{eq:ynrdw}, \eqref{eq:ynidw} and \eqref{eq:ynjkdw}. In the quaternionic case a term by term computation was still possible, for general Clifford algebras $Cl_{p,q}$ with $n > 9$ this is impossible even with current symbolic Clifford computer algebra systems (CAS), like the CLIFFORD package for MAPLE (\cite{RA:CLIFFORD}). 
\end{rmk}

The hypercomplex vector derivative of the hypercomplex FIR output signal \eqref{eq:ymAdw} allows us now to easily establish the hypercomplex vector derivative of the cost function $E_m$ as 
\be 
  E_m \partial_{\bv{w}} 
  = -2 \,[\sum_{A=1}^{2^n}e_{m,A} \,\phi'(s_{m,A}) \,\vect{e}_{A}]\, \widetilde{\bv{x}}
  \label{eq:ClEdwfin}
\ee 

Finally inserting \eqref{eq:ClEdwfin} into the weight update formula \eqref{eq:Clgraddesc} we obtain the split hypercomplex adaptive filtering algorithm (SHAFA) weight update as
\be 
  \bv{w}_{m+1} 
  = \bv{w}_{m} 
    + \mu \,[\sum_{A=1}^{2^n}e_{m,A} \,\phi'(s_{m,A}) \,\vect{e}_{A}]\, \widetilde{\bv{x}}.
  \label{eq:Clwtupdate}
\ee 
\begin{rmk}
Equations \eqref{eq:ClEdwfin} and \eqref{eq:Clwtupdate} are the consequent \textit{generalizations} of the quaternionic formulas \eqref{eq:Edwfin} and \eqref{eq:wtupdate}, respectively, to arbitrary hypercomplex algebras $Cl_{p,q}$.
\end{rmk}

\subsection{Convergence analysis for split hypercomplex adaptive filtering algorithm (SHAFA)}

The \textit{a posteriori} error for the SHAFA is
\be
  \epost_m = d_m -\Phi(\bv{w}_{m+1}\cdot\bv{x}_m),
\ee 
and the \textit{a priori} error 
\be 
  \eprior_m = d_m -\Phi(\bv{w}_{m}\cdot\bv{x}_m),
\ee 
Both are related by the first order \textit{Taylor series expansion}
\be 
  |\epost_m|^2 = |\eprior_m|^2 + \langle(|\eprior_m|^2\partial_{\bv{w}})\cdot\widetilde{\Delta\bv{w}_m}\rangle.
  \label{eq:Cltaylor}
\ee 
It involves the principal involution of the weight update
\begin{align}
  \widetilde{\Delta\bv{w}_m}
  = (\bv{w}_{m+1}-\bv{w}_m)^{\sim} 
  = \mu\, \bv{x} \sum_{A=1}^{2^n}\eprior_{m,A}\phi'_A(s_{m})\widetilde{\vect{e}_{A}}
  \label{eq:pinvwtupdate}
\end{align}
of \eqref{eq:Clwtupdate}, $\bv{x}=\bv{x}_m$, and the error gradient $|\eprior_m|^2\partial_{\bv{w}}$ of \eqref{eq:ClEdwfin}. For convergence of the SHAFA, we need
$|\epost_m|^2 < |\eprior_m|^2$ under the assumptions of small learning rates $\mu$.

Inserting \eqref{eq:ClEdwfin} and \eqref{eq:pinvwtupdate}, the second term of the Taylor expansion \eqref{eq:Cltaylor} gives
\begin{align}
  &\langle(|\eprior_m|^2\partial_{\bv{w}})\cdot\widetilde{\Delta\bv{w}_m}\rangle
  \nonumber \\
  &= -2\mu\,\langle[\sum_{A=1}^{2^n}\eprior_{m,A}\phi'_A(s_{m})\vect{e}_{A}]\, 
  (\widetilde{\bv{x}} \cdot \bv{x}) 
      \,[\sum_{B=1}^{2^n}\eprior_{m,B}\phi'_B(s_{m})\widetilde{\vect{e}_{B}}]\rangle
  \\
  &= -2\mu\sum_{A,B=1}^{2^n}\eprior_{m,A}\,\phi'_A(s_{m})\,\eprior_{m,B}\,\phi'_B(s_{m})
     \langle\, (\widetilde{\bv{x}} \cdot \bv{x}) 
      \,\widetilde{\vect{e}_{B}}\vect{e}_{A}\rangle,
  \label{eq:CLTaylor2}
\end{align}
where we used the symmetry of the scalar product for the second equality.

The Taylor expansion \eqref{eq:Cltaylor} can thus be expressed as
\be
  |\epost_m|^2 = |\eprior_m|^2 [1-M],
  \label{eq:Cleerel}
\ee
with
\begin{align} 
   M &= 2\mu\frac{1}{|\eprior_m|^2}\sum_{A,B=1}^{2^n}
        \eprior_{m,A}\,\phi'_A(s_{m}) \, \eprior_{m,B}\,\phi'_B(s_{m})
     \langle\, (\widetilde{\bv{x}} \cdot \bv{x}) 
      \,\widetilde{\vect{e}_{B}}\vect{e}_{A}\rangle
  \nonumber \\
  &= 2 \mu\frac{1}{|\eprior_m|^2} \langle (\widetilde{\bv{x}} \cdot \bv{x}) \,\widetilde{F}F\,\rangle
  = 2 \mu\frac{1}{|\eprior_m|^2} \langle F(\widetilde{\bv{x}} \cdot \bv{x}) \widetilde{F}\,\rangle
  \nonumber \\
  &= 2 \mu\frac{1}{|\eprior_m|^2} \sum_{l=1}^L \langle F \widetilde{x}_{m-l+1} x_{m-l+1}\widetilde{F}\,\rangle
  = 2 \mu\frac{1}{|\eprior_m|^2} \sum_{l=1}^L |F \widetilde{x}_{m-l+1}|^2,
  \label{eq:CldefM}
\end{align}
where we used the symmetry of the scalar product 
$ \langle (\widetilde{\bv{x}} \cdot \bv{x}) \widetilde{F}F \,\rangle
= \langle F (\widetilde{\bv{x}} \cdot \bv{x}) \widetilde{F}\,\rangle $,
expanded $\widetilde{\bv{x}} \cdot \bv{x}$ according to \eqref{eq:Clexpip}, 
used $(F\widetilde{x}_{m-l+1})^{\sim} = {x}_{m-l+1}\widetilde{F}$,  \eqref{eq:Clmodulus}, and defined 
\be 
  F = \sum_{A=1}^{2^n}\eprior_{m,A}\,\phi'_A(s_{m})\vect{e}_{A},
  \qquad
  |F|^2 = |\widetilde{F}F|
        = \sum_{A=1}^{2^n}\eprior_{m,A}^2\,(\phi'_A(s_{m}))^2.
  \label{eq:ClFdef}
\ee 
The last expression in \eqref{eq:CldefM} shows that 
$\langle (\widetilde{\bv{x}} \cdot \bv{x}) \,\widetilde{F}F\,\rangle
= |(\widetilde{\bv{x}} \cdot \bv{x}) \ast \widetilde{F}F|$. 
For convergence we need $0<M<1$ in \eqref{eq:Cleerel}. 
Using the hypercomplex Cauchy-Schwarz inequality of Theorem \ref{th:HCSine} for 
$|(\widetilde{\bv{x}} \cdot \bv{x}) \ast \widetilde{F}F|$,
we can estimate \eqref{eq:CldefM} for positive $\mu$ as
\be 
  0 < M = 2\mu \frac{1}{|\eprior_m|^2} |(\widetilde{\bv{x}} \cdot \bv{x}) \ast \widetilde{F}F|
    \leq 2\mu \frac{1}{|\eprior_m|^2} |\widetilde{\bv{x}} \cdot \bv{x}| |\widetilde{F}F| .
  \label{eq:Clmu1}
\ee 
In turn we can apply a $2^n$D Cauchy-Schwarz identity to $|F|^2$, i.e.
\be 
  |F|^2 
  = \sum_{A=1}^{2^n}\eprior_{m,A}^2\,(\phi'_A(s_{m}))^2
  \leq \sum_{A=1}^{2^n}\eprior_{m,A}^2\,\sum_{B=1}^{2^n}(\phi'_B(s_{m}))^2
  = |\eprior_m|^2 |P|^2
\ee 
with definition
$
  P = \sum_{B}^{2^n}\phi'_B(s_{m})\vect{e}_{B}
$.
Therefore we obtain for positive $\mu$ the estimate 
$
  0 < M < 2\mu |\widetilde{\bv{x}} \cdot \bv{x}| |P|^2, 
$
 and can ensure \textit{convergence} ($0<M<1$) by demanding 
\be 
  0 < \mu < \frac{1}{2 |\widetilde{\bv{x}} \cdot \bv{x}| |P|^2},
  \qquad 
  |P|^2 = \sum_{B=1}^{2^n}(\phi'_B(s_{m}))^2.
  \label{eq:Clmulimit}
\ee 

For \textit{scalar} $\widetilde{\bv{x}} \cdot \bv{x}$ we have 
\be 
  \widetilde{\bv{x}} \cdot \bv{x} 
  = \langle \widetilde{\bv{x}} \cdot \bv{x}\rangle 
  = \sum_{l=1}^L \langle \widetilde{x}_{m-l+1} x_{m-l+1}\rangle 
  = \sum_{l=1}^L | x_{m-l+1}|^2,
  \label{eq:scalarxx}
\ee 
and therefore
$
  M = 2\mu \frac{1}{|\eprior_m|^2} \langle \widetilde{\bv{x}} \cdot \bv{x}\rangle |F|^2
$.
The condition for convergence is then slightly modified to
\be 
  0 < \mu < \frac{1}{2 \langle \widetilde{\bv{x}} \cdot \bv{x}\rangle |P|^2}.
  \label{eq:Clmulimit_scxx}
\ee 
\begin{rmk}
  $\widetilde{\bv{x}} \cdot \bv{x}$ is \textit{scalar} for the algebras of complex numbers and quaternions, but \textit{not} in general.
For example, for $Cl_{1,0}=Cl_1$, with algebra basis $\{1, e_1\}$,  $\bv{x}=1+e_1=\widetilde{\bv{x}}$ gives the non-scalar $\widetilde{\bv{x}} \cdot \bv{x}=2(1+e_1)\notin \R$.
\end{rmk}

\section{Adaptive amplitude SHAFA (AASHAFA) \label{sc:AASHAFA}}

\subsection{Adaptive amplitude split hypercomplex adaptive filtering}

For hypercomplex real world signals with \textit{large dynamical ranges} we now construct a split hypercomplex adaptive FIR filter algorithm with trainable \textit{adaptive amplitudes} (AASHAFA).
We define componentwise
\be 
  \phi_A(s_m) = \lambda_{m,A} \varphi_A(s_m) = \lambda_{m,A} \varphi(s_{m,A}), 
  \quad 1 \leq A \leq 2^n,
  \quad s_m = \bv{w}_m \cdot \bv{x}_m,
\ee 
where $\lambda_{m,A}\in \R$ is the \textit{amplitude} for the $\vect{e}_{A}$ blade part of the hypercomplex number, and $\varphi: \R \rightarrow \R$, is the real nonlinearity with \textit{unit} amplitude applied to every blade part. 

The \textit{error} is defined as
\be 
  e_m = \sum_{A=1}^{2^n} e_{m,A}\vect{e}_{A}, \quad  
  e_{m,A} = d_{m,A} - \lambda_{m,A} \varphi_A(s_m),
  \quad 1 \leq A \leq 2^n.
  \label{eq:errormA}
\ee 
The \textit{cost function} is
\be 
  E_m = |e_m|^2 = \sum_{A=1}^{2^n}e_{m,A}^2.
  \label{eq:Emcost}
\ee 
The gradient based updates of the component amplitudes $\lambda_{m,A}$, with \textit{learning rate}\footnote{In principle it would be possible to optimize the learning further by introducing individual componentwise learning rates $\rho_A \in \R, 1\leq A \leq 2^n$.} $\rho \in \R$, are
\begin{align} 
  \lambda_{m+1,A} 
  &= \lambda_{m,A} - \frac{1}{2} \rho \frac{\partial E_m}{\partial_{\lambda_{m,A}}}
  = \lambda_{m,A} - \frac{1}{2} \rho \frac{\partial e_{m,A}^2}{\partial_{\lambda_{m,A}}}
  = \lambda_{m,A} - \rho e_{m,A} \frac{\partial e_{m,A}}{\partial_{\lambda_{m,A}}}
  \nonumber \\
  &= \lambda_{m,A} + \rho \,e_{m,A}\, \varphi_A(s_m),
\end{align} 
where we inserted $E_m$ of \eqref{eq:Emcost} for the second equality and $e_{m,A}$ of \eqref{eq:errormA} for the fourth equality.

\subsection{Convergence analysis for AASHAFA\label{sc:AASHconv}}

In \textit{adaptive amplitude} split hypercomplex adaptive filtering each amplitude parameter $\lambda_{m,A}, 1 \leq A \leq 2^n$, controls the nonlinearity in the $\vect{e}_{A}$ blade component dimension. We therefore investigate the convergence of each $\lambda_{m,A}$ \textit{separately}. 

The componentwise \textit{a priori} errors $\eprior_{m,A}$ and the \textit{a posteriori} errors
$\epost_{m,A}$, $1 \leq A \leq 2^n$, are
\be 
  \eprior_{m,A} = d_{m,A} - \lambda_{m,A} \varphi_A(\bv{w}_m \cdot \bv{x}_m),
  \quad
  \epost_{m,A} = d_{m,A} - \lambda_{m,A} \varphi_A(\bv{w}_{m+1} \cdot \bv{x}_m),
\ee 
respectively.
We consider the $A$-term of the Taylor series expansion \eqref{eq:Cleerel} corresponding to $\lambda_{m,A}$
\be 
  |\epost_{m,A}|^2 = |\eprior_{m,A}|^2 + \langle(|\eprior_{m,A}|^2\partial_{\bv{w}}) \cdot (\Delta_A\bv{w}_{m})^{\sim}\rangle,
  \label{eq:CltaylorA}
\ee 
where $\Delta_A\bv{w}_{m}$ is the weight update due to 
$E_{m,A} = |\eprior_{m,A}|^2 = \eprior_{m,A}^2$.

We now compute 
\be 
 \eprior_{m,A}\partial_{\bv{w}}
  = -\lambda_{m,A}\varphi'_A(s_{m}){\vect{e}_{A}}\widetilde{\bv{x}_m},
\ee 
just like in \eqref{eq:ymAdw}, using $d_{m,A}\partial_{\bv{w}}=0$, replacing $\phi(s_{m,A}) \rightarrow \lambda_{m,A}\varphi(s_{m,A})$ and defining $\varphi'_A(s_{m}) = \varphi'(s_{m,A})$.
This gives for the hypercomplex vector derivative of the cost function
\be 
  E_{m,A} \partial_{\bv{w}} 
  = |\eprior_{m,A}|^2\partial_{\bv{w}}
  = \eprior_{m,A}^2\partial_{\bv{w}}
  = -2\eprior_{m,A}\lambda_{m,A}\varphi'_A(s_{m}){\vect{e}_{A}}\widetilde{\bv{x}_m}.
  \label{eq:EmAdw}
\ee
We therefore get the \textit{weight update} 
\be 
  \Delta_A\bv{w}_{m}
  = -\frac{1}{2}\mu E_{m,A} \partial_{\bv{w}} 
  = \mu\,\eprior_{m,A}\,\lambda_{m,A} \,
    \varphi'_A(s_{m}){\vect{e}_{A}}\widetilde{\bv{x}_m}.
  \label{eq:DAwmup}
\ee 
Inserting \eqref{eq:EmAdw} and \eqref{eq:DAwmup} in \eqref{eq:CltaylorA} we obtain
\begin{align} 
  |\epost_{m,A}|^2 
  &= |\eprior_{m,A}|^2 
  -2 \mu \eprior_{m,A}^2 \lambda^2_{m,A} \varphi'^2_A(s_{m}) 
  \langle\widetilde{\bv{x}_m} \cdot \bv{x}_m \rangle
  \nonumber \\
  &= |\eprior_{m,A}|^2 \,[1-2\mu \lambda^2_{m,A} \varphi'^2_A(s_{m}) 
  \langle\widetilde{\bv{x}_m} \cdot \bv{x}_m \rangle],
  \label{eq:CltaylorAM}
\end{align} 
where we used 
$
  \langle\vect{e}_{A}(\widetilde{\bv{x}_m} \cdot \bv{x}_m) \widetilde{\vect{e}_{A}} \rangle
  = \langle(\widetilde{\bv{x}_m} \cdot \bv{x}_m) \widetilde{\vect{e}_{A}}\vect{e}_{A} \rangle
  \stackrel{\eqref{eq:Clbasis_scp}}{=} \langle\widetilde{\bv{x}_m} \cdot \bv{x}_m \rangle
$,
compare \eqref{eq:scalarxx}.

For \textit{convergence} we must therefore have in \eqref{eq:CltaylorAM} that
\be 
  0 < 1-2\mu \lambda^2_{m,A} \varphi'^2_A(s_{m}) 
  \langle\widetilde{\bv{x}_m} \cdot \bv{x}_m \rangle < 1.
\ee 
The adaptive amplitude parameters $\lambda_{m,A}, 1 \leq A \leq 2^n$, thus have the \textit{stability bounds} 
\be 
  0 < \lambda^2_{m,A} 
    < \frac{1}{2\mu \langle\widetilde{\bv{x}_m} \cdot \bv{x}_m \rangle \varphi'^2_A(s_{m}) },
  \qquad 1 \leq A \leq 2^n,
\ee
which explicitly \textit{depend} on the learning rate (step size) $\mu$.

\section{Conclusion}

In our present work we conducted an algebraically consequent quaternionic analysis of split quaternion adaptive filtering.
As results we obtain theoretical corrections of the algorithms as well as improved convergence, compared to \cite{UTM:SQNAF}.

We then extended the quaternionic approach
with the construction of a general split \textit{hypercomplex} adaptive FIR filtering algorithm (SHAFA), and for hypercomplex signals with large dynamic range we constructed 
an \textit{adapative amplitude} split hypercomplex adaptive FIR filtering algorithm (AASHAFA).
We investigated the SHAFA and AASHAFA learning algorithms and their convergence.
We thus established new algorithms based on a 
sound theoretical foundation in general Clifford algebras, with 
complex, hyperbolic number, and quaternionic split adaptive FIR filtering (optionally with adaptive amplitudes for large dynamic range signals) as special cases. 

We emphasize that this theoretical work is absolutely essential, since the high dimensional, non-commutative nature of hypercomplex numbers requires the consequent use of hypercomplex (multivector) algebra and hypercomplex differential calculus (\cite{HS:CAtoGC, RA:CLIFFORD} and \cite{EH:MVDC}), which are non-trival generalizations of real and complex mathematics. 
In this new framework 
an enormous range of applications to the processing of hypercomplex signals opens up, e.g. in geographic information systems (GIS) (see \cite{YL:CAUSTA}), meteorology (\cite{STTYF:Cyclone}), ocean currents, projective (homogeneous) and conformal geometric algebra (\cite{HL:IAGR, DFM:GAfCS, HTBY:Carrier} and \cite{EH:CPRSCGA}), electromagnetic signals (\cite{DL:GAP, EH:RPGA}), attractor prediction (\cite{SB:thesis}), and the like. 


\appendix
\section{Geometric interpretation of Clifford algebra\label{ap:geoint}}

The parts of grade $0$, $(s-r)$, $(r-s)$, and $(s+r)$, respectively, of the geometric product of an $r$-vector $A_r\in Cl_{p,q}$ with an $s$-vector $B_s\in Cl_{p,q}$ 
\begin{align}
\langle A_r B_s \rangle_{0} &= A_r \ast B_s, \quad
\langle A_r B_s \rangle_{s-r} = A_r \rfloor B_s, \nonumber \\
\langle A_r B_s \rangle_{r-s} &= A_r \lfloor B_s, \quad
\langle A_r B_s \rangle_{r+s} = A_r \wedge B_s,
\label{eq:gaprods}
\end{align} 
are called scalar product, left contraction, right contraction, and (associative) outer product, respectively, compare \cite{HL:IAGR}, \cite{DFM:GAfCS} and \cite{HTBY:Carrier}. These definitions extend by linearity to the corresponding products of general multivectors. The various derived products of \eqref{eq:gaprods} are related to each other, e.g. by
\be 
  (A\wedge B)\rfloor C = A\rfloor (B\rfloor C),
  \qquad
  \forall A,B,C \in Cl_{p,q}.
  \label{eq:ABCrel}
\ee 
 Note that for vectors $a,b$ in $\R^{p,q} \subset Cl_{p,q}$ we have
\be 
ab = a \rfloor b + a \wedge b, \quad
a \rfloor b = a \lfloor b = a \bullet b = a \ast b,
\ee 
where $a \bullet b$ is the inner product of $\R^{p,q}$. The geometric interpretation of the bivector $a \wedge b=-b \wedge a$ is an oriented parallelogram area in space with sense ($\pm$ sign). Higher order outer products (blades) $A_r=a_1\wedge \ldots \wedge a_r$ of $r$ linearly independent vectors $a_1, \ldots, a_r \in \R^{p,q}$, $1\leq r \leq n$, are interpreted as oriented $r$-dimensional parallelepipeds in space with orientation and sense. For non zero $A_r^2 = A_r \ast A_r \in \R\setminus \{0\}$, we can define the (right and left) inverse blade $A_r^{-1} = A_r/(A_r^2)$. For example every non-isotropic vector $b \in \R^{p,q}$, $\varepsilon_b|b|^2 = b^2 \neq 0$, $\varepsilon_b = \mathrm{sign}(b^2)$ has inverse $b^{-1} = b/(b^2)$.

The projection and rejection of vector $a$ onto (from) the non-isotropic vector $b$, are defined as
\begin{align}
  P_{b}(a) &= (a\rfloor \frac{b}{|b|})\frac{b}{\varepsilon_b|b|} = (a\rfloor b)b^{-1}, 
  \nonumber \\
  P_{b}^{\perp}(a) &= a - P_{b}(a) = (ab - a\rfloor b)b^{-1} = (a\wedge b)b^{-1},
\end{align}
respectively.
This can be generalized to projections and rejections of blades $A \in Cl_{p,q}$ onto (from) non-isotropic blades $B \in Cl_{p,q}$
\be 
  P_{B}(A) = (A\rfloor B)B^{-1}, 
  \qquad
  P_{B}^{\perp}(A) = (A\wedge B)B^{-1},
  \label{eq:proj}
\ee
respectively.

All vectors $b$ parallel to a non zero vector $a\in \R^{p,q}$ span a zero parallelogram area with $a$, i.e. the line space spanned by $a\in \R^{p,q}$ is given by $V(a) = \{b\in \R^{p,q}: b\wedge a = 0 \}$. Similarly a subspace of $\R^{p,q}$ spanned by $r$, $1\leq r \leq n$, linearly independent vectors $a_1, \ldots, a_r \in \R^{p,q}$,  is given by $V(a_1, \ldots, a_r) = \{b\in \R^{p,q}: b\wedge a_1\wedge \ldots \wedge a_r = 0 \}$. This subspace representation is called outer product null space representation (OPNS).

The \textit{duality} operation is defined as multiplication by the unit inverse pseudoscalar $I^{-1}=I/(I^2)$ (of maximum grade $n$) of the geometric algebra $Cl_{p,q}$. 
Given an $r$-dimensional subspace $V(a_1, \ldots, a_r)\in \R^{p,q}$ specified by its OPNS representation blade $A_r=a_1\wedge \ldots \wedge a_r$, then its \textit{dual} representation (as inner product null space [IPNS]) is given by the $(n-r)$-blade 
\be 
  A_r^{\ast} = A_r{I}^{-1} 
       = A_r\rfloor {I}^{-1} 
       = \langle A_r {I}^{-1}\rangle_{n-r}.
\ee 
The OPNS representation by $A_r$ and the dual IPNS representation by $A_r^{\ast}$ are directly related by duality  
\begin{align}
  \forall x\in \R^{p,q}:\,\,\,
  x \rfloor A_r^{\ast} 
  = x \rfloor (A_r\rfloor{I}^{-1})
  = (x \wedge A_r)\rfloor{I}^{-1} 
  = (x \wedge A_r){I}^{-1} ,
\end{align}
which holds again because of \eqref{eq:ABCrel}.
Therefore we have $\forall x \in \R^{p,q}$
\be 
   x \wedge A_r = 0
   \quad \Leftrightarrow \quad x \rfloor  A_r^{\ast} =0 .
\ee

\section{Proof of hypercomplex Cauchy-Schwarz type inequality \label{ap:HCSin}}

\noindent\textit{Proof.}
Assume two general multivectors $M,N \in Cl_{p,q}$ and a real parameter $t\in \R$. The following norm square will always be positive 
\begin{align} 
  0
  &\leq |M+tN|^2 
  = (M+tN) \ast (M+tN)^{\sim}
  \nonumber \\
  &= M\ast \widetilde{M} 
    + t(M\ast \widetilde{N}  + N\ast \widetilde{M})
    + t^2N\ast \widetilde{N}. 
  \label{eq:CSproof1}
\end{align} 
Because scalars are invariant under the principal involution, we must have
\be 
  M\ast \widetilde{N} = (M\ast \widetilde{N})^{\sim} = N\ast \widetilde{M} 
  \quad \Rightarrow \quad
  M\ast \widetilde{N} + N\ast \widetilde{M} 
  = 2 M\ast \widetilde{N}.  
\ee 
Equation \eqref{eq:CSproof1} can thus be simplified to 
\be 
  0\leq |M+tN|^2 = |M|^2 + 2t \,M\ast \widetilde{N} + t^2 |N|^2.
\ee 
For $|M|^2 + 2t \,M\ast \widetilde{N} + t^2 |N|^2$ to be always positive, the following discriminant $D$ must be negative
\be 
  D = (2 M\ast \widetilde{N})^2 - 4 |M|^2 |N|^2 \leq 0.
\ee 
We conclude that
\begin{gather} 
  D\leq 0 \quad \Leftrightarrow \quad
  (M\ast \widetilde{N})^2 \leq |M|^2 |N|^2
  \quad\Leftrightarrow \quad
  |M\ast \widetilde{N}| \leq |M| |N|.
\end{gather} 
If we finally replace $N \rightarrow \widetilde{N}$ and use $|N| = |\widetilde{N}|$ we get
\be 
  |M\ast N| \leq |M| |N|.
\ee 

\noindent
QED.


\end{document}